\documentclass[conference]{IEEEtran}

\pdfoutput=1
\usepackage{graphicx}

\usepackage{cite}
\usepackage{amsmath,amssymb,amsfonts}
\usepackage{algorithm}
\usepackage{booktabs}
\usepackage{algorithmic}
\usepackage{graphicx}
\usepackage[font=small]{caption}
\usepackage{tabularx}
\usepackage{subfig}
\usepackage{pgffor}
\usepackage{hyperref}
\usepackage{soul}
\usepackage{array, makecell}
\usepackage{cellspace}
\usepackage{tikz}
\hypersetup{draft}

\newcommand{\group}[1]{ \left( #1 \right) }

\newcommand{\vect}[1]{ {\boldsymbol{#1}} }

\DeclareMathOperator*{\argmin}{arg\,min}

\newcommand{\func}[2]{ { #1 \group{#2} } }

\newcommand{\dataset}{{\cal D}}

\newcommand{\vI}{\vect{I}}

\newcommand{\vQ}{\vect{Q}}

\newcommand{\vx}{\vect{x}}

\newcommand{\vy}{\vect{y}}

\newcommand{\loss}{\mathcal{L}}
\newcommand{\set}[1]{{\left\{#1\right\}}}

\DeclareMathOperator{\argmax}{argmax}

\begin{document}

\title{An Adversarial Approach for Explainable AI in Intrusion Detection
       Systems}

\author{\IEEEauthorblockN{ Daniel L. Marino, Chathurika S. Wickramasinghe,
    Milos Manic }
	\IEEEauthorblockA{\textit{Department of Computer Science} \\
	\textit{Virginia Commonwealth University}\\
	Richmond, USA \\
	marinodl@vcu.edu, misko@ieee.org}}

\maketitle

\begin{abstract}
Despite the growing popularity of modern machine learning techniques
  (e.g. Deep Neural Networks) in cyber-security applications,
  most of these models are perceived as a black-box for the user.
Adversarial machine learning offers an approach to increase our understanding
  of these models.
In this paper we present an approach to generate explanations for
  incorrect classifications made by data-driven Intrusion Detection Systems (IDSs)
An adversarial approach is used to find the minimum
  modifications (of the input features) required to correctly
  classify a given set of misclassified samples.
The magnitude of such modifications is used to visualize the most relevant
  features that explain the reason for the misclassification.
The presented methodology generated satisfactory explanations that describe the
  reasoning behind the mis-classifications, with descriptions that match
  expert knowledge.
The advantages of the presented methodology are:
  1) applicable to any classifier with defined gradients.
  2) does not require any modification of the classifier model.
  3) can be extended to perform further diagnosis
    (e.g. vulnerability assessment) and gain further understanding of the system.
Experimental evaluation was conducted on the NSL-KDD99 benchmark dataset
  using Linear and Multilayer perceptron classifiers.
The results are shown using intuitive visualizations in order to improve
  the interpretability of the results.

\end{abstract}


\begin{tikzpicture}[overlay, remember picture]
\path (current page.north west) ++(1.5,-0.3) node[below right,text width=19cm] {
 \small
 \copyright 2018  IEEE.  Personal  use  of  this  material  is  permitted.
 Permission  from  IEEE  must  be  obtained  for  all  other  uses,  in  any current
 or  future  media,  including reprinting/republishing this material for
 advertising or promotional purposes, creating new collective works,
 for resale or redistribution to servers or lists, or reuse of any copyrighted
 component of this work in other works.};
\end{tikzpicture}
\begin{tikzpicture}[overlay, remember picture]
\path (current page.south west) ++(1.7,2.3) node[below right,text width=9.1cm] {
  \footnotesize
  \noindent\rule{9cm}{0.4pt}\\
  Accepted version of the paper appearing in the
  proceedings of the 44th Annual Conference of the
  IEEE Industrial Electronics Society, IECON 2018.
  };
\end{tikzpicture}

\section{Introduction}
\label{section:introduction}
The increasing incorporation of Cyber-based methodologies for monitoring and
  control of physical systems has made critical infrastructure
  vulnerable to various cyber-attacks such as: interception,
  removal or replacement of information, penetration of unauthorized users and
  viruses \cite{Kwon2017} \cite{Sridhar2012} \cite{Raj2010} \cite{Alvaro}.
Intrusion detection systems (IDSs) are an essential tool
  to detect such malicious attacks \cite{Lee1999}.

Extendibility and adaptability are essential requirements for an IDS \cite{Buczak2015}.
Every day, new strains of Cyber-attacks are created with the objective of deceiving
  these systems.
As a result, machine learning and data-driven intrusion detection systems
  are increasingly being used in cyber-security applications \cite{Buczak2015}.
Machine learning techniques such as deep neural
  networks have been successfully used in IDSs \cite{Buczak2015}
  \cite{kasun2018towards}.
However, these techniques often work as a black-box model for the user.
  \cite{DBLP:journals/corr/abs-1708-08296} \cite{kasun2018towards}.

\begin{figure}[t]
\centering
    \includegraphics[scale=0.4]{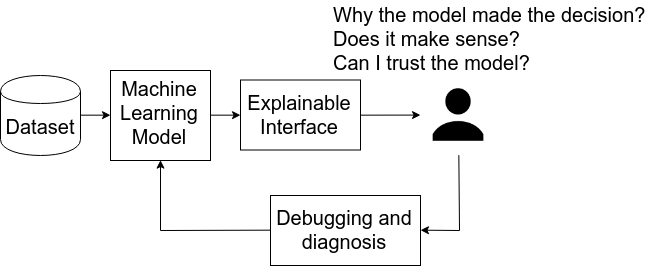}
\caption{Explainable AI \cite{gunning2017explainable} is concerned with
  designing of interfaces that help users to understand the decisions made
  by a trained machine learning model.}
\label{fig:explainable_ai}
\end{figure}

\begin{figure*}[t]
\centering
    \includegraphics[scale=0.5]{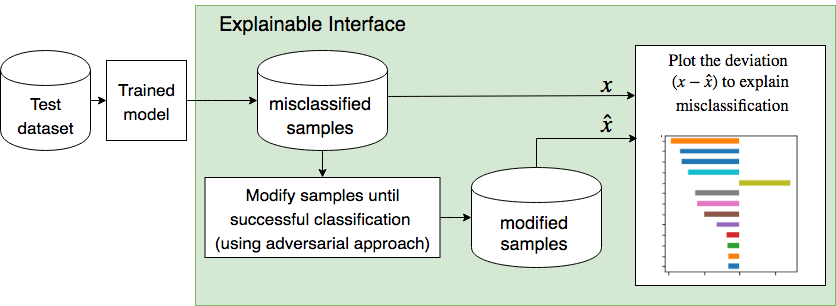}
\caption{Overview of the presented explainable interface.
The approach provides explanations for misclassified samples.
An adversarial approach is followed to modify the misclassified samples until
  the model assigns the correct class.
This approach provides a mechanism to understand the decision boundaries of the
  classifier.}
\label{fig:overview}
\end{figure*}

With machine learning being increasingly applied in the operation of critical
  systems, understanding the reason behind the decisions made by a model
  has become a common requirement to the point that
  governments are starting to include it into legislation
  \cite{eu2016european} \cite{goodman2017european}.
Recent developments in the machine learning community
  have been focused in the development of methods which are more interpretable
  for the users.
Explainable AI (Figure \ref{fig:explainable_ai}) \cite{ DBLP:journals/corr/abs-1708-08296}
  makes use of visualizations and natural language descriptions to explain
  the reasoning behind the decisions made by the machine learning model.

It is crucial that the inner workings of data-driven models are transparent for the
  engineers designing IDSs.
Decisions presented by explainable models can be easily interpreted by a human,
  simplifying the process of knowledge discovery.
Explainable approaches help on diagnosing, debugging, and understanding the
  decisions made by the model, ultimately increasing the trust on the
  data-driven IDS.

In the case of data-driven IDS, when the model is presented with new attacks
  where data is not available, the model might mis-classify an attack as normal,
  leading to a breach in the system.
Understanding the reasons behind the misclassification of particular samples
  is the first step for debugging and diagnosing the system.
Providing clear explanations for the cause of misclassification is essential
  to decide which steps to follow in order to prevent future attacks.

In this paper, we present an explainable AI interface for
  diagnosing data-driven IDSs.
We present a methodology to explain incorrect classifications made by the model
  following an adversarial approach.

Although adversarial machine learning is usually used to deceive the classifier,
in this paper we use it to generate explanations by finding the
  minimum modifications required in order to correctly classify the
  misclassified samples.
The difference between the original and modified samples provide information
  of the relevant features responsible for the misclassification.
We show the explanations provide satisfactory insights behind the
  reasoning made by the data-driven models, being congruent with the
  expert knowledge of the task.

The rest of the paper is organized as follows:
  Section \ref{section:adversarial_machine_learning} presents an overview of
    adversarial machine learning;
  Section \ref{section:methodology} describes the presented methodology for
    explainable IDS systems.
  Section \ref{section:results} describes the experimental results carried out
    using the NSL-KDD99 intrusion detection benchmark dataset.
  Section \ref{section:conclusion} concludes the paper.

\section{Adversarial Machine Learning}
\label{section:adversarial_machine_learning}
\newcommand{\class}{k}
\newcommand{\adv}{{\hat{\vx}}}
\newcommand{\real}{{\vx_0}}
\newcommand{\model}[1]{\func{p}{\vy | #1, w}}
\newcommand{\modelk}[1]{\func{p}{\vy = \class | #1, w}}
\newcommand{\normalize}[1]{\func{g}{#1}}
\newcommand{\nadv}{\hat{\vx}}
\newcommand{\nreal}{{\vx_0}}
\newcommand{\target}{{\hat{\vy}}}

Adversarial machine learning has been extensively used in cyber-security
  research to find vulnerabilities in data-driven models
  \cite{Lowd2005} \cite{Barreno2010} \cite{Wagner2002}
  \cite{DBLP:journals/corr/LiMSRJ17}
  \cite{barreno2006can} \cite{frederickson2018attack}.
Recently, there has been an increasing interest on adversarial samples given
  the susceptibility of Deep Learning models to these type of attacks
  \cite{goodfellow6572explaining} \cite{kurakin2016adversarial}.

Adversarial samples are samples crafted in order to change the output of a
  model by making small modifications into a reference (usually real)
  sample \cite{goodfellow6572explaining}.
These samples are used to detect blind spots on ML algorithms.

Adversarial samples are crafted from an attacker perspective to
  evade detection, confuse the classifier \cite{goodfellow6572explaining},
  degrade performance \cite{papernot2017practical} and/or
  gain information about the model
  or the dataset used to train the model \cite{frederickson2018attack}.
Adversarial samples are also useful from a defender point of view given that
  they can be used to perform vulnerability assessment \cite{barreno2006can},
  study the robustness against noise, improve generalization and
  debug the machine learning model \cite{papernot2017practical}.

In general, the problem of crafting an adversarial sample is stated as follows
  \cite{frederickson2018attack}:
\begin{align}
  \max_\adv \quad & \loss(\adv) - \Omega(\adv) \label{eq:general_attack}\\
  s.t.      \quad & \adv \in \phi(\vx)            \nonumber
\end{align}
where:
\begin{itemize}
  \item $\loss(\vx)$ measures the impact of the adversarial sample in the model,
  \item $\Omega(\vx)$ measures the capability of the defender to detect the
    adversarial sample.
  \item $\adv \in \phi(\vx)$ ensures the crafted sample $\adv$ is inside
    the domain of valid inputs. It also represents the capabilities of the
    adversary.
\end{itemize}

The objective of Eq. (\ref{eq:general_attack}) can be interpreted as crafting an
  attack point $\adv$ that maximizes the impact of the attack, while
  minimizing the chances of the attack to be detected.
The definition of $\loss(\vx)$ will depend on the intent of the attack and
  the available information about the model under attack.
For example:
  1) $\loss(\vx)$ can be a function that measures the difference between
    a target class $\target$ and the output from the model;
  2) $\Omega(\vx)$ measures the discrepancy between the reference $\real$ and
    the modified sample $\adv$

In this paper, instead of deceiving the classifier, we use Equation
  \ref{eq:general_attack} to find the minimum number of modifications needed
  to correctly classify a misclassified sample.
The methodology is described in detail in section \ref{section:methodology}.

\section{Explaining misclassifications using adversarial machine learning}
\label{section:methodology}

In this paper we are interested in generating explanations for incorrect
  estimations made by a trained classifier.
Figure \ref{fig:overview} presents an overview of the presented explainable
  interface.
The presented methodology modifies a set of misclassified samples until they
  are correctly classified.
The modifications are made following an adversarial approach: finding the
  minimum modifications required to change the output of the model.
The difference between the modified samples and the original real samples
  is used to explain the output from the classifier, illustrating the
  most relevant features that lead to the misclassification.

In the following sections we explain in detail each component of the
  presented explainable interface.

\subsection{Data-driven classifier}

The classifier $\func{p}{y=\class | \vx, w}$ estimates the probability of a given sample
  $\vx$ to belong to class $\class$.
The classifier is parameterized by a set of parameters $w$ that are learned
  from data.
Learning is performed using standard supervised learning.
Given a training dataset $\dataset=\set{(\vx^{(i)}, \vy^{(i)})}_i^{M}$ of
  $M$ samples, the parameters $w$ of the model are obtained by minimizing the
  cross-entropy:
\begin{align*}
  w^* = \argmin_{w} \sum_i^M \func{H}{\vy^{(i)}, \model{\vx^{(i)}}}
\end{align*}
where $\vy^{(i)}$ is a one-hot encoding representation of the class:
\begin{align*}
  \vy_i =
  \begin{cases}
    1 \quad \text{if $\vx^{(i)} \in$ class  $i$} \\
    0 \quad \text{otherwise}
  \end{cases}
\end{align*}

Depending on the complexity of the model $\model{\vx}$ and the dataset
  $\dataset$, the model may misclassify some of the samples.
We are interested on provide explanations for these incorrect estimations.

\subsection{Modifying misclassified samples}

In this paper, instead of deceiving the classifier, we make use of adversarial
  machine learning to understand why some of the samples are being mis-classified.
The idea of this approach is that adversarial machine learning can help us to
  understand the decision boundaries of the learned model.

The objective is to find the minimum modifications needed in order to change
  the output of the classifier for a real sample $\real$.
This is achieved by finding an adversarial sample $\adv$ that is classified as
  $\target$ while minimizing the distance between the real sample ($\real$)
  and the modified sample $\adv$:

\begin{align}
  \min_{\adv}  \quad &
    \group{\nadv - \nreal}^T \vQ \group{\nadv - \nreal}
    \label{eq:adversarial_program} \\
   \text{s.t}  \quad &
    \argmax_{\class}{\modelk{\nadv}} = \target
    \label{eq:classification_constraint}\\
                     & \vx_{\min} \preceq \adv \preceq \vx_{\max}
                       \nonumber
\end{align}
where $\vQ$ is a symmetric positive definite matrix, that allows the user to
  specify a weight in the quadratic difference metric.

The program in Equation \ref{eq:adversarial_program} can serve multiple
  purposes depending on how $\real$ and $\target$ are specified.
For the purpose of explaining incorrect classifications, in this paper,
  $\real$ represents the real misclassified samples that serve
  as a reference for the modified samples $\adv$.
Furthermore, the value of $\target$ is set to the correct class of $\real$.
In this way, the program (\ref{eq:adversarial_program}) finds a modified
  sample $\adv$ that is as close as possible to the real misclassified
  sample $\real$, while being correctly classified by the model as $\target$.

We constrain the sample $\adv$ to be inside the bounds
  $(\vx_{min}, \vx_{max})$.
These bounds are extracted from the maximum and minimum values found in the
  training dataset.
This constraint ensures that the adversarial example is inside the domain
  of the data distribution.

The class $\target$ of the adversarial sample is specified by the user.
Note that $\vy \neq \target$, i.e. the class of the real sample
  is different from the class of the adversarial sample.

The optimization problem in Eq. \ref{eq:adversarial_program} provides a clear
  objective to solve.
However, the problem as stated in Eq. \ref{eq:adversarial_program} is not
  straightforward to solve using available deep-learning
  optimization frameworks.
In order to simplify the implementation, we modify the way the constraints
  are satisfied by moving the constraint in Eq.
  \ref{eq:classification_constraint} into the objective function:

\begin{align}
  \min_{\adv}  \quad
      & H(\target, \model{\nadv}) \alpha \vI_{(\adv, \target)}
      \label{eq:final_adversarial_program}\\
      & \quad + \group{\nadv - \nreal}^T \vQ \group{\nadv - \nreal}
      \nonumber \\
   \text{s.t}  \quad
      & \vx_{\min} \preceq \adv \preceq \vx_{\max}
      \nonumber
\end{align}

where:
\begin{itemize}
  \item $(\real, \vy)$ is a reference sample from the dataset
  \item $\adv$ is the modified version of $\real$ that makes the estimation
    of the class change from $\vy$ to the target $\target$
  \item $H(\target, \model{\nadv})$ is the cross-entropy between the
    estimated adversarial sample class $\model{\nadv}$
    and the target class $\target$
  \item $\vI_{(\adv, \target)}$ is an indicator function that specifies whether
    the adversarial sample $\adv$ is being classified as $\target$
    \begin{align*}
      \vI_{(\adv, \target)} =
      \begin{cases}
        0 \quad \text{if $\argmax_k \modelk{\nadv} = \target$} \\
        1 \quad \text{otherwise}
      \end{cases}
    \end{align*}
    This function provides a mechanism to stop the modifications once the
      sample $\adv$ is classified as $\target$.
    We assume this function is not continuous, hence, the gradients with
      respect the inputs are not required.
  \item $\alpha$ is a scale factor that can be used to weight the
    contribution of the cross-entropy $H$ to the objective loss.
\end{itemize}

The problem stated in Eq. \ref{eq:final_adversarial_program} can be seen as
  an instance of the adversarial problem stated in Eq. \ref{eq:general_attack},
  where the cross-entropy $H$ represents the effectiveness ($\loss$)
  of the modifications while the quadratic difference represents
  the discrepancy ($\Omega$) between $\real$ and $\adv$.

\begin{figure*}[t]
\centering
  \subfloat[misclassified samples $\real$]{
    \includegraphics[trim={0 0 7.5cm 0},clip,scale=0.7]{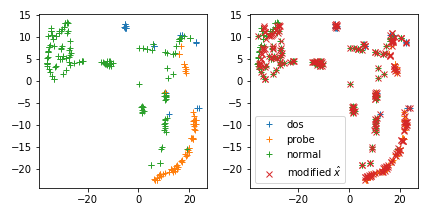}
    \label{fig:tsne_corrected_mlp_a}
  } \quad \quad
  \subfloat[misclassified $\real$ and modified samples $\adv$]{
    \includegraphics[trim={8.0cm 0 0 0},clip,scale=0.7]{{tsne_corrected_mlp_constrained}.png}
    \label{fig:tsne_corrected_mlp_b}
  }
\caption{
  t-SNE visualization of misclassified samples and corresponding modified
    samples for the MLP classifier.
  The legend shows the class to which the misclassified samples belong to.
  The axes correspond to an embedded 2D representation of the samples.
  The figure shows there is no clear distinction between the misclassified real
    samples $\real$ and the modified samples $\adv$, demonstrating that
    small modifications can be used to change the output of the classifier.
}
\label{fig:tsne_corrected_mlp}
\end{figure*}

\subsection{Explaining incorrect estimations}
We used the adversarial approach stated in Eq. \ref{eq:final_adversarial_program},
  to generate an explanation for incorrect classification.
Using a set of $\real$ misclassified samples as reference,
  we use Eq. \ref{eq:final_adversarial_program} to find the minimum modifications
  needed to correctly classify the samples.
We use as target $\target$ the real class of the samples.

The explanations are generated by visualizing the difference $(\real - \adv)$
  between the misclassified samples $\real$ and the modified samples $\adv$.
This difference shows the deviation of the real features $\real$ from what the model
  considers as the target class $\target$.

The explanations can be generated for individual samples $\real$ or for
  a set of misclassified samples.
We used the average deviation $(\real - \adv)$ to present the explanations
  for a set of misclassified samples.

\section{Experiments and Results}
\label{section:results}
\subsection{Dataset}
For experimental evaluation, we used the NSL-KDD intrusion detection
  dataset \cite{tavallaee2009detailed}.
The NSL-KDD dataset is a revised version of the KDD99 dataset
  \cite{KDD99}, a widely
  used benchmark dataset used for intrusion detection algorithms.
The NSL-KDD dataset removes redundant and duplicate records found in the
  KDD99 dataset, alleviating some of the problems of the KDD99 dataset
  mentioned in \cite{tavallaee2009detailed}.

The NSL-KDD consists of a series of aggregated records extracted from a
  packet analyzer log.
Besides normal communications, the dataset contains records of attacks that
  fall in four main categories: DOS, R2L, U2R and proving.
For our experiments, we only considered normal, DOS and probe classes.

The dataset consists of 124926 training samples and 16557 testing samples.
We used the same split provided by the authors of the dataset.
To alleviate the effects of the unbalanced distribution of the samples over the
  classes, we trained the models extracting mini-batches with an equal ratio of
  samples from each class. Samples were extracted with replacement.
A detailed description of the dataset features can be found in
  \cite{perona2013gurekddcup}.

The dataset samples were normalized in order to make the quadratic distance
  metric in Equation \ref{eq:final_adversarial_program} invariant to the
  features scales.
We used the mean and standard deviation of the training dataset for
  normalization:
\begin{align*}
  \vx^{(i)} \leftarrow
    \dfrac{ \vx^{(i)} - \func{\text{MEAN}}{\set{\vx|\vx \in \dataset}}}
                        {\func{\text{STD}}{\set{\vx|\vx \in \dataset}}}
\end{align*}

\subsection{Classifiers}

One of the advantages of the methodology presented in this paper is that
  it works for any classifier which has a defined gradient $\nabla_{\vx}H$
  of the cross-entropy loss with respect to the inputs.

For the experimental section, we used the following data-driven models:
  1) a linear classifier, and
  2) a Multi Layer Perceptron (MLP) classifier with ReLU activation function.
We used weight decay (L2 norm) and early-stopping regularization for
  training both models.

Table \ref{table:accuracy} shows the accuracy achieved with the linear and the
  MLP classifiers. We can see that the MLP classifier provides higher accuracy
  in the training and testing datasets.

\begin{table}[h]
  \captionsetup{justification=centering}
  \caption{Accuracy of Classifiers}
  \centering
	\begin{tabular}{l|l|l}
		\toprule
      classifier & train & test      \\
		\midrule
    Linear &  0.957  &  0.936   \\
		MLP    &  0.995  &  0.955   \\
		\bottomrule
	\end{tabular}
	\label{table:accuracy}
\end{table}

\subsection{Modified misclassified samples}
We used t-SNE \cite{maaten2008visualizing} to visualize the
  misclassified samples ($\real$) and the modified/corrected samples
  ($\adv$) found using Equation \ref{eq:final_adversarial_program}.
t-SNE is a dimensionality reduction technique commonly used to
  visualize high-dimensional datasets.

Figure \ref{fig:tsne_corrected_mlp} shows the visualization of the
  misclassified samples $\real$ and the corresponding modified samples $\adv$
  using t-SNE.
This figure shows the effectiveness of the presented methodology to find
  the minimum modifications needed to correct the output of the classifier.
No visual difference in the visualization can be observed between the real
  and the modified samples.
The modified samples $\adv$ are close enough to the real samples $\real$ that
  the modified samples occlude the real samples in Figure \ref{fig:tsne_corrected_mlp_b}.

\subsection{Explaining incorrect estimations}

\begin{figure}[t]
\centering
  \subfloat[Normal samples misclassified as DOS using Linear model]{
    \includegraphics[trim={0 0 0 0.5cm},clip,scale=0.7]{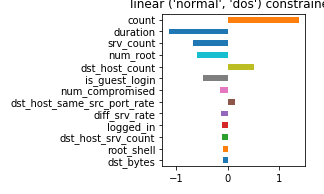}
    \label{fig:corrected_deviation_linear(normal,dos)_constrained}
  } \\
  \subfloat[Normal samples misclassified as DOS using MLP model]{
    \includegraphics[trim={0 0 0 0.5cm},clip,scale=0.65]{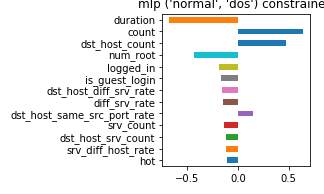}
    \label{fig:corrected_deviation_mlp(normal,dos)_constrained}
  }
\caption{
  Explanation for Normal samples being misclassified as DOS using the difference
  between real samples $\real$ and modified samples $\adv$.
}
\label{fig:corrected_deviation(normal,dos)_constrained}
\end{figure}

\begin{figure}[t]
\centering
  \includegraphics[trim={0 0 0 1.2cm},clip,scale=0.6]{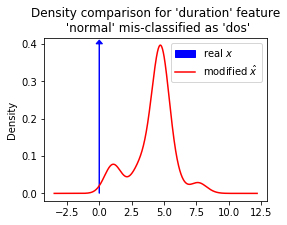}
\caption{
  Comparison of \textit{duration} feature between $\real$ (Normal samples
    misclassified as DOS) and  $\adv$ (modified samples).
  The model considers connections with zero duration suspicious.
  The modified samples have an increased duration in order to correctly classify
    the samples as Normal.
}
\label{fig:duration_distribution}
\end{figure}

\begin{figure}[t]
\centering
  \subfloat[DOS samples misclassified as Normal using Linear model]{
    \includegraphics[trim={0 0 0 0.5cm},clip,scale=0.7]{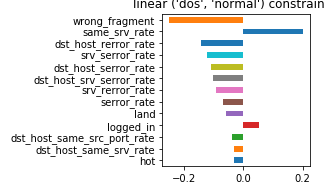}
    \label{fig:corrected_deviation_linear(dos,normal)_constrained}
  } \\
  \subfloat[DOS samples misclassified as Normal using MLP model]{
    \includegraphics[trim={0 0 0 0.5cm},clip,scale=0.7]{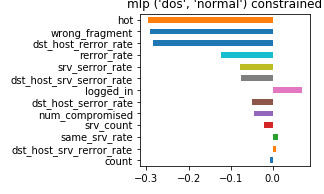}
    \label{fig:corrected_deviation_mlp(dos,normal)_constrained}
  }
\caption{
  Explanation for DOS samples being misclassified as Normal using the difference
  between real samples $\real$ and modified samples $\adv$.
}
\label{fig:corrected_deviation(dos,normal)_constrained}
\end{figure}

\begin{figure*}[t]
\centering
  \subfloat[Deviation for continuous features]{
    \includegraphics[trim={0 0 0 0.5cm},clip,scale=0.73]{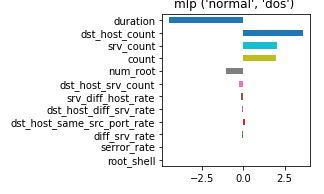}
    \label{fig:corrected_deviation_mlp(normal,dos)}
  }
  \subfloat[Comparison of categorical features distributions]{
    \includegraphics[trim={0 0 7.0cm 0},clip,scale=0.7]{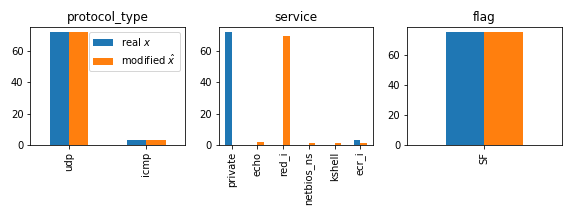}
    \label{fig:corrected_histogram_mlp(normal,dos)}
  }
\caption{Explanation for Normal samples mis-classified as DOS
  while taking into account categorical features (MLP model).}
\label{fig:corrected_mlp(normal,dos)}
\end{figure*}

Figure \ref{fig:corrected_deviation(normal,dos)_constrained} shows the generated explanation
  for Normal samples being incorrectly classified as DOS.
Figure \ref{fig:corrected_deviation_linear(normal,dos)_constrained}
  shows the explanations for the Linear model while
  Figure \ref{fig:corrected_deviation_mlp(normal,dos)_constrained} shows the
  explanations for the MLP model.

We observed that the explanations for both models provide similar qualitative
  explanations.
Figures \ref{fig:corrected_deviation_linear(normal,dos)_constrained}
  and \ref{fig:corrected_deviation_mlp(normal,dos)_constrained}
  can be naturally interpreted as follows:

{\centering
\textit{Normal samples were mis-classified as DOS because:}}
\begin{itemize}
  \item \textit{high} number of connections to the same host (count) and to the
    same destination address (dst\_host\_count)
  \item \textit{low} connection duration (duration)
  \item \textit{low} number of operations performed as root in the connection
    (num\_root)
  \item \textit{low} percentage of samples have successfully logged in
    (logged\_in, is\_guest\_login)
  \item \textit{high} percentage of connections originated from the
    same source port (dst\_host\_same\_src\_port\_rate)
  \item \textit{low} percentage of connections directed to different
    services (diff\_srv\_rate)
  \item \textit{low} number of connections were directed to the same
    destination port (dst\_host\_srv\_count)
\end{itemize}

Figures \ref{fig:corrected_deviation_linear(normal,dos)_constrained}
  and \ref{fig:corrected_deviation_mlp(normal,dos)_constrained} that a high
  number of connections with low duration and low login success rate
  is responsible for misclassifying Normal samples as DOS.
These attributes are clearly suspicious and match what a
  human expert would consider as typical behavior of a DOS attack,
  providing a satisfactory explanation for the incorrect classification.

A more detailed view of the difference between the \textit{duration} of
  real ($\real$) and modified ($\adv$) samples is presented in
  Figure \ref{fig:duration_distribution}.
This figure shows that all misclassified Normal samples had connections with
  a duration of zero seconds, which the classifier considers suspicious for
  a Normal behavior.

The graphs also provide a natural way to extract knowledge and understand the
  concepts learned by the model.
For example, figures \ref{fig:corrected_deviation_linear(normal,dos)_constrained}
  and \ref{fig:corrected_deviation_mlp(normal,dos)_constrained}
  show the classifier considers suspicious when there is
  a low percentage of connections directed to different services
  (low \textit{diff\_srv\_rate}).

A parallel analysis can be performed to other misclassified samples.
Figure \ref{fig:corrected_deviation(dos,normal)_constrained}
  provides explanations for DOS attacks misclassified as Normal connections
  for Linear and MLP models.
Overall, Figures \ref{fig:corrected_deviation_linear(dos,normal)_constrained}
  and \ref{fig:corrected_deviation_mlp(dos,normal)_constrained},
  show that the samples are misclassified as Normal because they have:
  (a) lower error rate during the connection
  (b) higher login success.
These features are usually expected to belong to normal connections, which
  successfully explains the reason for the misclassification.

The explanations shown in Figures
  \ref{fig:corrected_deviation(normal,dos)_constrained} and
  \ref{fig:corrected_deviation(dos,normal)_constrained}
  do not take categorical features into account.
In order to include categorical features into the analysis, we perform a
  round operation to the inputs of the indicator $\vI_{(\adv, \target)}$.
\footnote{Given that we do not use the gradient of $\vI_{(\adv, \target)}$
  during the optimization process, we are allowed to
  include discontinuous operations like the rounding function}
This ensures the objective function in Equation \ref{eq:final_adversarial_program}
  takes into consideration the effects of the rounding operation.

Figure \ref{fig:corrected_mlp(normal,dos)} shows the explanations generated
  when considering categorical features.
Figure \ref{fig:corrected_deviation_mlp(normal,dos)} shows the deviation of
  continuous features, providing the same information as the explanations from
  Figures \ref{fig:corrected_deviation(normal,dos)_constrained} and
  \ref{fig:corrected_deviation(dos,normal)_constrained}.
Figure \ref{fig:corrected_histogram_mlp(normal,dos)} shows the comparison
  between the histograms of misclassified samples and modified samples.
Figure \ref{fig:corrected_histogram_mlp(normal,dos)} shows that
  \textit{protocol\_type} was not modified, suggesting that this
  feature is not relevant in order to explain the misclassification.
On the other hand, the \textit{service} feature was modified in almost all
  samples.
The figure shows that most of the Normal samples misclassified as DOS
  used a \textit{private} service.
Changing the service value helps the classifier to correctly estimate the class
  of the samples.
The explanation shows that the model considers communication with
  \textit{private} service as suspicious.

\section{Conclusion}
\label{section:conclusion}
In this paper, we presented an approach for generating explanations
  for the incorrect classification of a set of samples.
The methodology was tested using an Intrusion Detection benchmark dataset.

The methodology uses an adversarial approach to find the minimum modifications
  needed in order to correctly classify the misclassified samples.
The modifications are used to find and visualize the relevant features
  responsible for the misclassification.
Experiments were performed using Linear and Multilayer perceptron classifiers.
The explanations were presented using intuitive plots that can be easily
  interpreted by the user.

The proposed methodology
  provided insightful and satisfactory explanations for the
  misclassification of samples, with results that match expert knowledge.
The relevant features found by the presented approach showed
  that misclassification often occurred on samples with conflicting
  characteristics between classes.
For example, normal connections with low duration and low login success are
  misclassified as attacks, while attack connections with low error rate and
  higher login success are misclassified as normal.

\section{Discussion}
\label{section:discussion}
An advantage of the presented approach is that it
  can be used for any differentiable model and
  any classification task.
No modifications of the model are required.
The presented approach only requires the gradients of the model cross-entropy
  w.r.t. the inputs.
Non-continuous functions can also be incorporated into the approach, for
  example rounding operations for integer and categorical features.

The presented adversarial approach can be extended to perform other analysis
  of the model.
For example, instead of finding modifications for correct the predicted class,
  the modifications can be used to deceive the classifier, which can be used
  for vulnerability assessment.
Future research will be conducted to incorporate the explanations to improve
  the accuracy of the model without having to include new data into the
  training procedure.

\vskip 0.2in
\bibliographystyle{IEEEtran}
\bibliography{my_bibliography}

\end{document}